\documentclass[letterpaper, 10 pt, conference]{ieeeconf}  

\IEEEoverridecommandlockouts                              

\usepackage{hyperref}

\usepackage{graphics} 
\usepackage{epsfig} 
\usepackage{times} 
\usepackage{amsmath} 
\usepackage{multirow}
\usepackage{booktabs}       

\title{\LARGE \bf
SemGauss-SLAM: Dense Semantic Gaussian Splatting SLAM
}

\author{Siting Zhu$^{1}$, Renjie Qin$^{1}$, Guangming Wang$^{2}$, Jiuming Liu$^{1}$, Hesheng Wang$^{1}$
\thanks{*This work was supported in part by the Natural Science Foundation of China under Grant 62225309, U24A20278, 62361166632, U21A20480 and 62403311. Corresponding Author: Hesheng Wang ({\tt\small wanghesheng@sjtu.edu.cn}).}
\thanks{$^{1}$School of Automation and Intelligent Sensing, Key Laboratory of System Control and Information Processing, Ministry of Education of China, Shanghai Jiao Tong University, Shanghai.}%
\thanks{$^{2}$ University of Cambridge, United Kingdom.}
}

\begin{document}

\maketitle
\thispagestyle{empty}
\pagestyle{empty}

\begin{abstract}

We propose SemGauss-SLAM, a dense semantic SLAM system utilizing 3D Gaussian representation, that enables accurate 3D semantic mapping, robust camera tracking, and high-quality rendering simultaneously. In this system, we incorporate semantic feature embedding into 3D Gaussian representation, which effectively encodes semantic information within the spatial layout of the environment for precise semantic scene representation. Furthermore,  we propose feature-level loss for updating 3D Gaussian representation, enabling higher-level guidance for 3D Gaussian optimization.
In addition, to reduce cumulative drift in tracking and improve semantic reconstruction accuracy, we introduce semantic-informed bundle adjustment. By leveraging multi-frame semantic associations, this strategy enables joint optimization of 3D Gaussian representation and camera poses, resulting in low-drift tracking and accurate semantic mapping.
Our SemGauss-SLAM demonstrates superior performance over existing radiance field-based SLAM methods in terms of mapping and tracking accuracy on Replica and ScanNet datasets, while also showing excellent capabilities in high-precision semantic segmentation and dense semantic mapping. Code will be available at \href{https://github.com/IRMVLab/SemGauss-SLAM}{https://github.com/IRMVLab/SemGauss-SLAM}. 

\end{abstract}

\section{Introduction}
\label{sec:intro}

Dense semantic Simultaneous Localization and Mapping (SLAM) is a fundamental challenge for robotic systems~\cite{rosinol2020kimera, mccormac2017semanticfusion} and autonomous driving~\cite{bao2022semantic,lianos2018vso}. It integrates semantic understanding of the environment into dense map reconstruction and performs pose estimation simultaneously. Traditional dense semantic SLAM  has limitations including its inability to predict unknown areas~\cite{liu2023unsupervised}. Subsequent semantic SLAM based on Neural Radiance Fields (NeRF)~\cite{mildenhall2021nerf} methods~\cite{zhu2024sni, li2023dns, haghighi2023neural} have addressed these drawbacks, but suffer from inefficient per-pixel raycasting rendering for real-time optimization of implicit scene representation and low-quality novel view semantic representation.

Recently, a novel radiance field based on 3D Gaussian Splatting (3DGS)~\cite{kerbl20233d} has demonstrated remarkable capability in scene representation, enabling high-quality and efficient rendering via splatting. Following the advantages of 3D Gaussian representation, 3DGS-based SLAM methods~\cite{yan2023gs, keetha2023splatam, yugay2023gaussian, matsuki2023gaussian, huang2023photo} have been developed to achieve photo-realistic mapping. However, most existing 3DGS SLAM systems focus on visual mapping to obtain RGB maps, where color information alone is insufficient for downstream tasks such as navigation. Moreover, current semantic NeRF-based SLAM methods suffer from cumulative drift in tracking, leading to degraded SLAM accuracy. Additionally, these methods struggle to achieve accurate semantic segmentation from novel viewpoints, indicating limited precision in online-reconstructed 3D semantic representation. Therefore, developing a low-drift and high-precision semantic SLAM system based on radiance field is essential and challenging.

\begin{figure}
  \centering
  \includegraphics[width=\linewidth]{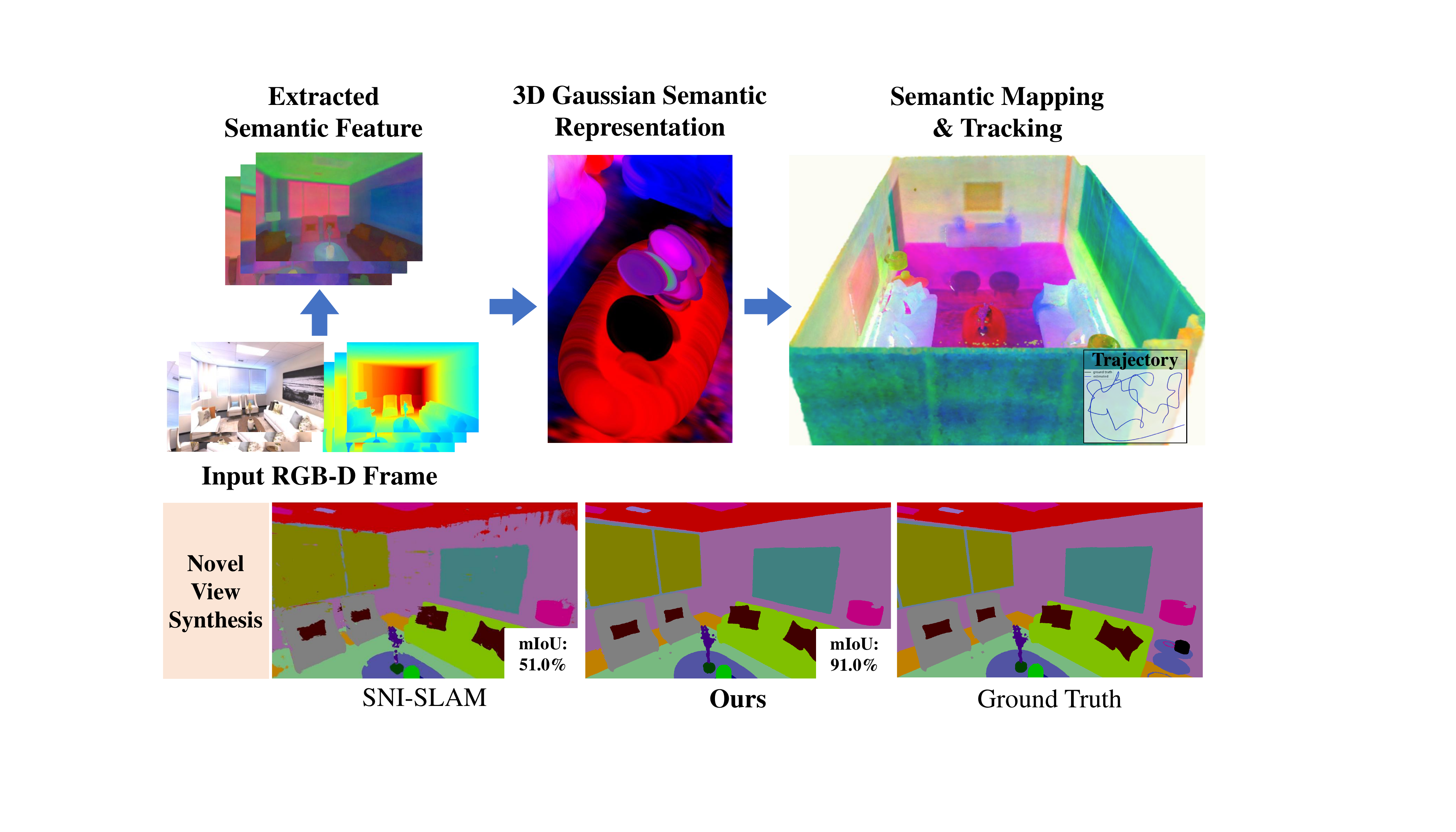}
  \vspace{-0.25in}
      \caption{Our SemGauss-SLAM incorporates semantic feature embedding into 3D Gaussian representation to perform dense semantic SLAM. 
    This modeling strategy not only achieves accurate semantic mapping, but also enables high-precision semantic novel view synthesis compared with other radiance field-based semantic SLAM. We visualize 3D Gaussian blobs with semantic embedding, showing the spatial layout of semantic Gaussian representation.
    Moreover, semantic mapping is visualized using semantic feature embedding, showing 3D semantic modeling of the scene.
    }
    \vspace{-0.1in}
\end{figure}


To overcome this challenge, we introduce a novel dense semantic SLAM system based on 3D Gaussian Splatting with semantic-informed bundle adjustment, which enables high accuracy in tracking, dense mapping, and semantic segmentation from novel view. Different from concurrent work SGS-SLAM~\cite{li2024sgs}, which uses color for semantic representation to achieve 3DGS semantic SLAM, our method integrates semantic feature embedding into 3D Gaussian for semantic modeling, resulting in a more discriminative and comprehensive understanding of the scene. Moreover, by leveraging the explicit 3D structure of 3D Gaussian representation, our embedding-based semantic Gaussian representation is capable of capturing the spatial distribution of semantic information, achieving high-accuracy novel view semantic segmentation. Furthermore, to reduce accumulated drift in radiance field-based semantic SLAM, we leverage semantic associations among co-visible frames and propose semantic-informed bundle adjustment (BA) for joint optimization of camera poses and 3D Gaussian representation. This design exploits the consistency of multi-view semantics for establishing constraints, which enables the reduction of cumulative drift in tracking and enhanced semantic mapping precision. Overall, we provide the following contributions:

\begin{itemize}
 \item We present SemGauss-SLAM, a dense semantic SLAM system based on 3D Gaussian, which can achieve accurate semantic mapping, photo-realistic reconstruction, and robust tracking. We incorporate semantic feature embedding into 3D Gaussian to achieve precise semantic scene representation. Moreover, feature-level loss is introduced to provide higher-level guidance for 3D Gaussian optimization, leading to high-precision scene optimization results.
 \item We perform semantic-informed bundle adjustment by leveraging multi-view semantic constraints for joint optimization of camera poses and 3D Gaussian representation, achieving low-drift tracking and accurate semantic mapping.
 \item We conduct extensive evaluations on challenging datasets, to demonstrate our method achieves superior performance compared with existing radiance field-based SLAM in mapping, tracking, semantic segmentation, and novel-view synthesis.
\end{itemize}
\section{Related Works}
\noindent\textbf{Traditional Semantic SLAM.}\hspace*{5pt}  
Traditional semantic SLAM employs explicit 3D representation such as surfels~\cite{mccormac2017semanticfusion}, mesh~\cite{rosinol2020kimera, tian2022kimera, hughes2022hydra} and Truncated Signed Distance Fields (TSDF)~\cite{schmid2022panoptic, mccormac2018fusion++, grinvald2019volumetric} for dense semantic mapping. SemanticFusion~\cite{mccormac2017semanticfusion} uses surfel representation and employs Conditional Random Field (CRF) for updating class probability distribution incrementally. Fusion++~\cite{mccormac2018fusion++} performs object-level SLAM where each object is reconstructed within its own TSDF volume and segmented based on estimated foreground probability. Kimera~\cite{rosinol2020kimera} utilizes visual-inertial odometry for pose estimation and generates dense semantic mesh maps. 
However, these explicit scene modeling methods not only require high storage space but also fail to achieve high-fidelity and complete 3D reconstruction.

\noindent\textbf{Neural Implicit SLAM.}\hspace*{5pt} 
Neural implicit representation~\cite{mildenhall2021nerf} has shown promising capability in scene reconstruction for dense mapping. iMAP~\cite{imap} first achieves mapping and tracking utilizing a single MLP network for scene representation. To overcome over-smoothed scene reconstruction and improve scalability, NICE-SLAM~\cite{zhu2022nice} adopts hierarchical feature grid representation. Following this, several works~\cite{yang2022vox, eslam, wang2023co, sandstrom2023point, kong2023vmap, hu2024cp} introduce more efficient scene representation, such as hash-based feature grid and feature plane, to achieve more accurate SLAM performance. 

For neural implicit semantic SLAM, DNS SLAM~\cite{li2023dns} utilizes 2D semantic priors and integrates multi-view geometry constraints for semantic reconstruction. SNI-SLAM~\cite{zhu2024sni} introduces feature collaboration and one-way correlation decoder for improved scene representation in semantic mapping. However, these methods require specific scene bounds for mapping and suffer from cumulative drift in pose estimation. 
In this paper, we leverage the explicit structure of 3D Gaussian for unbounded mapping and perform semantic-informed bundle adjustment utilizing multi-frame semantic constraints for conducting low-drift and high-quality dense semantic SLAM.

\noindent\textbf{3D Gaussian Splatting SLAM.}\hspace*{5pt} 
3D Gaussian Splatting~\cite{kerbl20233d} emerges as a promising 3D scene representation using a set of 3D Gaussians with learnable properties, including 3D center position, anisotropic covariance, opacity, and color. This representation is capable of quick differential rendering through splatting and
has a wide range of applications in dynamic scene modeling~\cite{yan2024street, zhou2023drivinggaussian}, semantic segmentation~\cite{ye2023gaussian, zhou2023feature} and scene editing~\cite{zhuang2024tip}.

Our main focus is on 3D Gaussian SLAM~\cite{yan2023gs, keetha2023splatam, matsuki2023gaussian, yugay2023gaussian}. These works emerge concurrently and all perform dense visual SLAM by leveraging scene geometry and appearance modeling capabilities of 3D Gaussian representation. SplaTAM~\cite{keetha2023splatam} introduces silhouette-guided optimization to facilitate structured map expansion
for dense mapping of visual SLAM. Gaussian Splatting SLAM~\cite{matsuki2023gaussian} performs novel Gaussian insertion and pruning for monocular SLAM. However, the capability of 3D Gaussian representation extends well beyond appearance and geometry modeling of the scene, as it can be augmented to perform semantic scene understanding. 
The concurrent work SGS-SLAM~\cite{li2024sgs} introduces semantic color associated with the Gaussian for semantic representation. However, this color-based semantic modeling method ignores the higher-level information inherent in semantics, which is insufficient for semantic representation of the environment. To address this limitation, we incorporate semantic feature embedding into 3D Gaussian for 3D semantic scene modeling to achieve high-precision dense semantic SLAM. 
\begin{figure*}
  \centering
  \includegraphics[width=\linewidth]{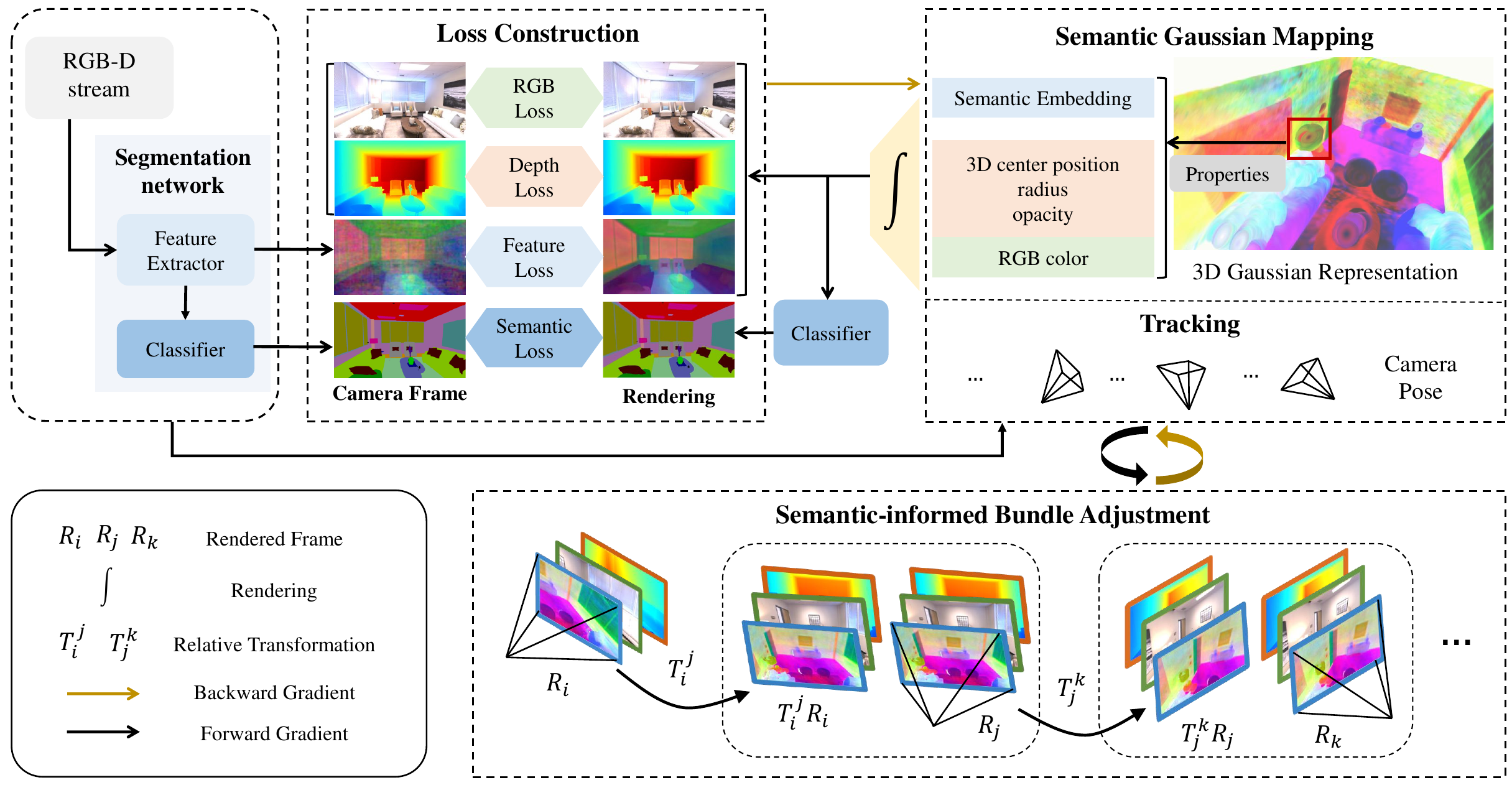}
  \vspace{-0.2in}
    \caption{\textbf{An overview of SemGauss-SLAM.} Our method takes an RGB-D stream as input. RGB images are fed into feature extractor to obtain semantic features. These features are then categorized by a pretrained classifier to attain semantic labels. Then, semantic features, semantic labels, along with the input RGB and depth data serve as supervision signals. In the meantime, semantic features and input RGB-D data propagate to 3D Gaussian blobs as initial properties of Gaussian representation. Rendered semantic features, RGB, and depth are obtained from 3D Gaussian splatting, while rendered semantic labels are attained by classifying rendered features. Supervision and rendered information are utilized for loss construction to optimize camera poses and 3D Gaussian representation. During the SLAM process, we utilize semantic-informed bundle adjustment based on multi-frame constraints for joint optimization of poses and 3D Gaussian representation.
    }
    \vspace{-0.1in}
  \label{fig:pipline}
\end{figure*}

\section{Method}
The overview of SemGauss-SLAM is shown in Fig.~\ref{fig:pipline}. 
Sec.~\ref{subsec:semantic mapping} introduces semantic Gaussian representation for dense semantic mapping, as well as the process of 3DGS-based semantic SLAM. Sec.~\ref{subsec:Loss Functions} introduces loss functions in the mapping and tracking process. Sec.~\ref{subsec:Bundle Adjustment} presents semantic-informed bundle adjustment to reduce accumulated drift in tracking and improve mapping accuracy.

\subsection{3D Gaussian Semantic Mapping and Tracking}
\label{subsec:semantic mapping}
\noindent\textbf{Semantic Gaussian Representation.}\hspace*{5pt}
We utilize a set of isotropy Gaussians with specific properties for scene representation in SLAM. To achieve semantic Gaussian mapping, we introduce a new parameter, semantic feature embedding, to each Gaussian for semantic representation. This design enables compact and efficient Gaussian semantic representation to capture the spatial semantic information of the environment. Moreover, it is crucial for 3D Gaussian representation, augmented by semantic feature embedding, to converge rapidly during optimization process for real-time mapping.
Therefore, instead of initializing these feature embeddings randomly, we propagate 2D semantic features extracted from images to 3D Gaussian as the initial values to achieve faster convergence of semantic Gaussian optimization. In this work, we use a universal feature extractor DINOv2~\cite{oquab2023dinov2}, followed by a pretrained classifier to construct the segmentation network. Overall, each Gaussian includes 3D center position \(\mu\), radius \(r\), color \(c=(r,g,b)\), opacity \(\alpha\), 16-channel semantic feature embedding \(e\), and is defined as standard Gaussian equation multiplied by opacity \(\alpha\):
\begin{equation}
 g(x) = \alpha \exp\left(-\frac{\|x-\mu\|^2}{2r^2}\right).
 \label{eq:opacity}
\end{equation}

\noindent\textbf{3D Gaussian Rendering.}\hspace*{5pt} Following Gaussian splatting~\cite{kerbl20233d}, we project 3D Gaussians to 2D splats for high-fidelity differentiable rendering, which facilitates explicit gradient flow for Gaussian scene optimization and pose estimation.
Specifically, for splatting of semantic feature embedding, we first sort all Gaussians in order of depth from near to far and then blend \(N\) ordered points projecting to pixel \(p=(u,v)\), obtaining 2D semantic feature maps \(E(p)\):
\begin{equation}
    E(p) = \sum_{i \in N} e_i g_i(p) \prod_{j=1}^{i-1} (1 - g_j(p)),
\label{eq:rgb render}
\end{equation}
where \(g_i(p)\) is computed as shown in Eq.(\ref{eq:opacity}) with \(\mu\) and \(r\) representing the rendered 2D Gaussians in pixel plane: 
\begin{equation}
\mu_{\text{pix}} = K_c\frac{T_k\mu}{d}, \ r_{\text{pix}} = \frac{fr}{d}, \ \text{where}~d = (T_k \mu)_z.
\end{equation}
\(K_c\) is calibrated camera intrinsic, \(T_k\) is estimated camera pose at frame \(k\), \(f\) is known focal length, \(d\) is the depth of the \(i\)-th Gaussian in camera coordinates.

For RGB and depth rendering, we follow the similar approach in Eq.(\ref{eq:rgb render}):
\begin{equation}
\begin{aligned}
C(p) = \sum_{i \in N} c_i g_i(p) \prod_{j=1}^{i-1} (1 - g_j(p)) ,\\
D(p) = \sum_{i \in N} d_i g_i(p) \prod_{j=1}^{i-1} (1 - g_j(p))
\end{aligned}
\end{equation}
where \(C(p)\) and \(D(p)\) represents splatted 2D color and depth images respectively. Moreover, given the camera pose, visibility information of Gaussian is required for mapping and tracking process, such as adding new Gaussian and loss construction. Therefore, following ~\cite{keetha2023splatam}, we render a silhouette image to determine visibility:
\begin{equation}
    Sil(p) = \sum_{i \in N} g_i(p) \prod_{j=1}^{i-1} (1 - g_j(p)).
\end{equation}

\noindent\textbf{Tracking Process.}\hspace*{5pt}
During tracking process, we keep 3D Gaussian parameters fixed and only optimize camera pose of current frame. Since the proximity between adjacent frames is relatively small, we assume a constant velocity model to obtain an initial pose estimation for the current frame. The camera pose is then iteratively refined through optimization by minimizing the loss between two components: differentiable rendering from 3D Gaussian representation and camera observation within the visible silhouette. 

\noindent\textbf{Mapping Process.}\hspace*{5pt}
Our system performs RGB mapping and semantic mapping simultaneously. Mapping process begins with the initialization of the scene representation, which is achieved by inverse transforming all pixels of the first frame to 3D coordinates and obtaining the initial 3D Gaussian representation. Then, when the overlap of coming frame with rendering of the existing map is less than half, we add a new Gaussian for incremental mapping. 

\subsection{Loss Functions}
\label{subsec:Loss Functions}
For optimization of semantic scene representation, we utilize cross-entropy loss for constructing semantic loss \(\mathcal{L}_s\). Furthermore, instead of solely using semantic loss for optimization, we introduce a feature-level loss \(\mathcal{L}_f\) for higher-level guidance of semantic optimization:
\begin{equation}
\mathcal{L}_{f}=\sum_{p \in P_M}|F_e-E(p)|,
\end{equation}
where $F_e$ is extracted features generated by DINOv2~\cite{oquab2023dinov2}-based feature extractor. \(P_M\) represents a set of all pixels in the rendered image. Compared with semantic loss, feature loss offers explicit supervision on intermediate features for semantic Gaussian optimization, leading to more robust and accurate semantic understanding of the scene.

We employ RGB loss \(\mathcal{L}_c\) and depth loss \(\mathcal{L}_d\) for optimizing scene color and geometry representation. \(\mathcal{L}_c\) and \(\mathcal{L}_d\) are both L1 loss constructed by comparing the RGB and depth splats with the input RGB-D frame. These loss functions are then utilized for mapping and pose estimation.

In the mapping process, we construct loss over all rendered pixels for 3D Gaussian optimization. Moreover, we add SSIM term to RGB loss following \cite{kerbl20233d}. The complete loss function for mapping is a weighted sum of the above losses:
\begin{equation}
\footnotesize
\mathcal{L}_{\text{mapping}} =  \sum_{p \in P_M}{(\lambda_{f_m} \mathcal{L}_{f}(p) +\lambda_{s_m} \mathcal{L}_{s}(p) + \lambda_{c_m} \mathcal{L}_{c}(p)+ \lambda_{d_m} \mathcal{L}_{d}(p))}.
\end{equation}
 \(\lambda_{f_m}\), \(\lambda_{s_m}\), \(\lambda_{c_m}\), \(\lambda_{d_m}\) are weighting coefficients.

In the tracking process, using an overly constrained loss function for pose estimation can lead to decreased accuracy in camera pose and increased processing time. Therefore, loss function for tracking is constructed based only on a weighted sum of RGB loss and depth loss:
\begin{equation}
\mathcal{L}_{\text{tracking}} =  \sum_{p \in P_T}{(\lambda_{c_t} \mathcal{L}_{c}(p)+ \lambda_{d_t} \mathcal{L}_{d}(p))},
\end{equation}
where \(\lambda_{c_t}\), \(\lambda_{d_t}\) are weighting coefficients in tracking process. \(P_T\) represents pixels that are rendered from well-optimized part of 3D Gaussian map, which is area that rendered visibility silhouette \(Sil(p)\) is greater than 0.99.
\subsection{Semantic-informed Bundle Adjustment}
\label{subsec:Bundle Adjustment}
Currently, existing radiance field-based semantic SLAM systems utilize the latest input RGB-D frame to construct RGB and depth loss for pose estimation. Subsequently, the scene representation of these SLAM systems is optimized using the estimated pose and the latest frame.
However, relying solely on single-frame constraint for pose optimization can lead to cumulative drift in the tracking process due to the absence of global constraints. Furthermore, using only single-frame information to optimize scene representation can result in globally inconsistent updates to the scene on the semantic level. 
To address this problem, we propose semantic-informed bundle adjustment to achieve joint optimization of 3D Gaussian representation and camera poses by leveraging multi-view constraints and semantic associations. 

In semantic-informed BA, we leverage the consistency of multi-view semantics to establish constraints. Specifically, rendered semantic feature is warped to its co-visible frame \(j\) using estimated relative pose transformation \(T_i^j\), and constructs loss with rendered semantic feature $\mathcal{G}(T_j, e)$ of frame \(j\) to obtain \(\mathcal{L}_{\text{BA-sem}}\): 
\begin{equation}
\mathcal{L}_{\text{BA-sem}} =\sum_{i=1}^{N-1}\sum_{j=i+1}^{N}{(|T_i^j \cdot \mathcal{G}(T_i, e) - \mathcal{G}(T_j, e)|)},
\label{eq:ba_sem}
\end{equation}
where \(\mathcal{G}(T_i, e)\) represents splatted semantic embedding from 3D Gaussian \(\mathcal{G}\) using camera pose \(T_i\). 
Moreover, to achieve geometry and appearance consistency, we also warp rendered RGB and depth to co-visible frames to construct loss with corresponding frames following similar approach in Eq.(\ref{eq:ba_sem}):
\begin{align}
\mathcal{L}_{\text{BA-rgb}} &= \sum_{i=1}^{N-1}\sum_{j=i+1}^{N}{(|T_i^j \cdot \mathcal{G}(T_i, c) - \mathcal{G}(T_j, c)|)}, \notag \\
\mathcal{L}_{\text{BA-depth}} &= \sum_{i=1}^{N-1}\sum_{j=i+1}^{N}{(|T_i^j \cdot \mathcal{G}(T_i, d) - \mathcal{G}(T_j, d)|)},
\end{align}
where \(\mathcal{G}(T_i, c)\) and \(\mathcal{G}(T_i, d)\) represents rendered RGB and depth of frame $i$.
Therefore, overall loss function \(\mathcal{L}_{\text{BA}}\) for joint optimization of corresponding frame poses and 3D Gaussian scene representation is the weighted sum of the above losses:
\begin{equation}
\mathcal{L}_{\text{BA}} = \lambda_{e} \mathcal{L}_{\text{BA-sem}} + \lambda_{c} \mathcal{L}_{\text{BA-rgb}} + \lambda_{d} \mathcal{L}_{\text{BA-depth}},
\end{equation}
where \(\lambda_{e}\), \(\lambda_{c}\), \(\lambda_{d}\) are weighting coefficients. 
This design leverages the fast rendering capability of 3D Gaussian to achieve joint optimization of scene representation and camera poses. Furthermore, semantic-informed BA integrates consistency and correlation of multi-perspective semantic, geometry, and appearance information, resulting in low-drift tracking and consistent mapping.

\section{Experiments}
\label{sec:Experiments}
\subsection{Experimental Setup}
\label{subsec:Experimental Setup}
\noindent\textbf{Datasets.}\hspace*{5pt}  
We evaluate the performance of SemGauss-SLAM on two datasets with semantic ground truth annotations, including 8 scenes on simulated dataset Replica~\cite{straub2019replica} and 5 scenes on real-world dataset ScanNet~\cite{dai2017scannet}.

\noindent\textbf{Metrics.}\hspace*{5pt}  
We follow metrics from~\cite{keetha2023splatam} to evaluate the SLAM system accuracy and rendering quality. For reconstruction metric, we use \textit{Depth L1 (cm)}. For tracking accuracy evaluation, we use \textit{ATE RMSE (cm)}~\cite{sturm2012benchmark}. Moreover, we use \textit{PSNR (dB)}, \textit{SSIM}, and \textit{LPIPS} for evaluating RGB image rendering performance. Semantic segmentation is evaluated with respect to \textit{mIoU (\%)} metric.

\begin{figure*}[tb]
  \centering
  \includegraphics[width=\linewidth]{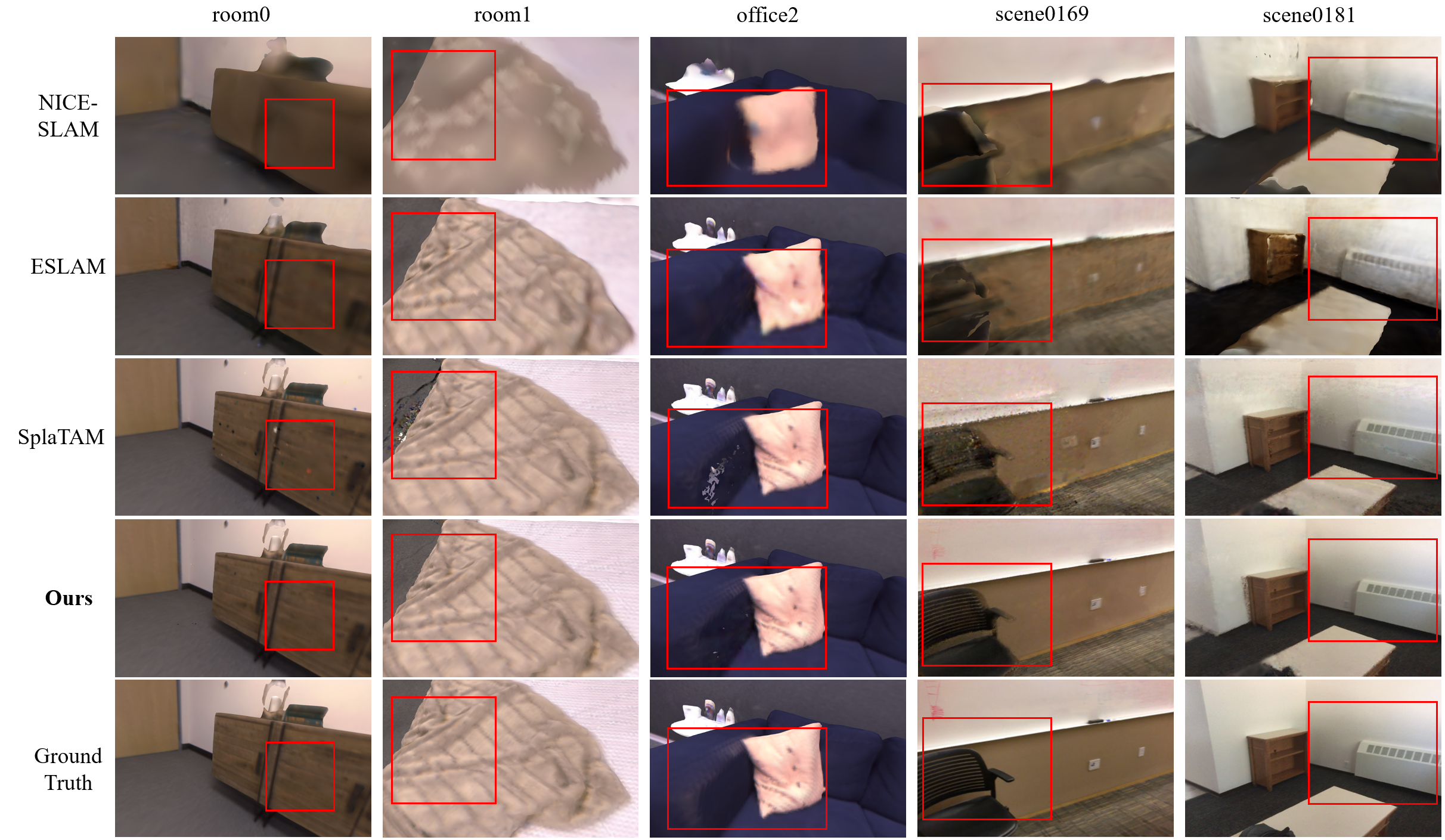}
  \vspace{-0.25in}
   \caption{Qualitative comparison on rendering quality of our method and baseline. We visualize 5 selected scenes of Replica ~\cite{straub2019replica} and ScanNet~\cite{dai2017scannet} dataset. Details are highlighted with red color boxes. Our method achieves photo-realistic rendering quality and higher completion of reconstruction, especially in areas with rich textural information.}
    \vspace{-0.1in}
  \label{fig:details}
\end{figure*}

\begin{figure}[t]
  \centering
  \includegraphics[width=\linewidth]{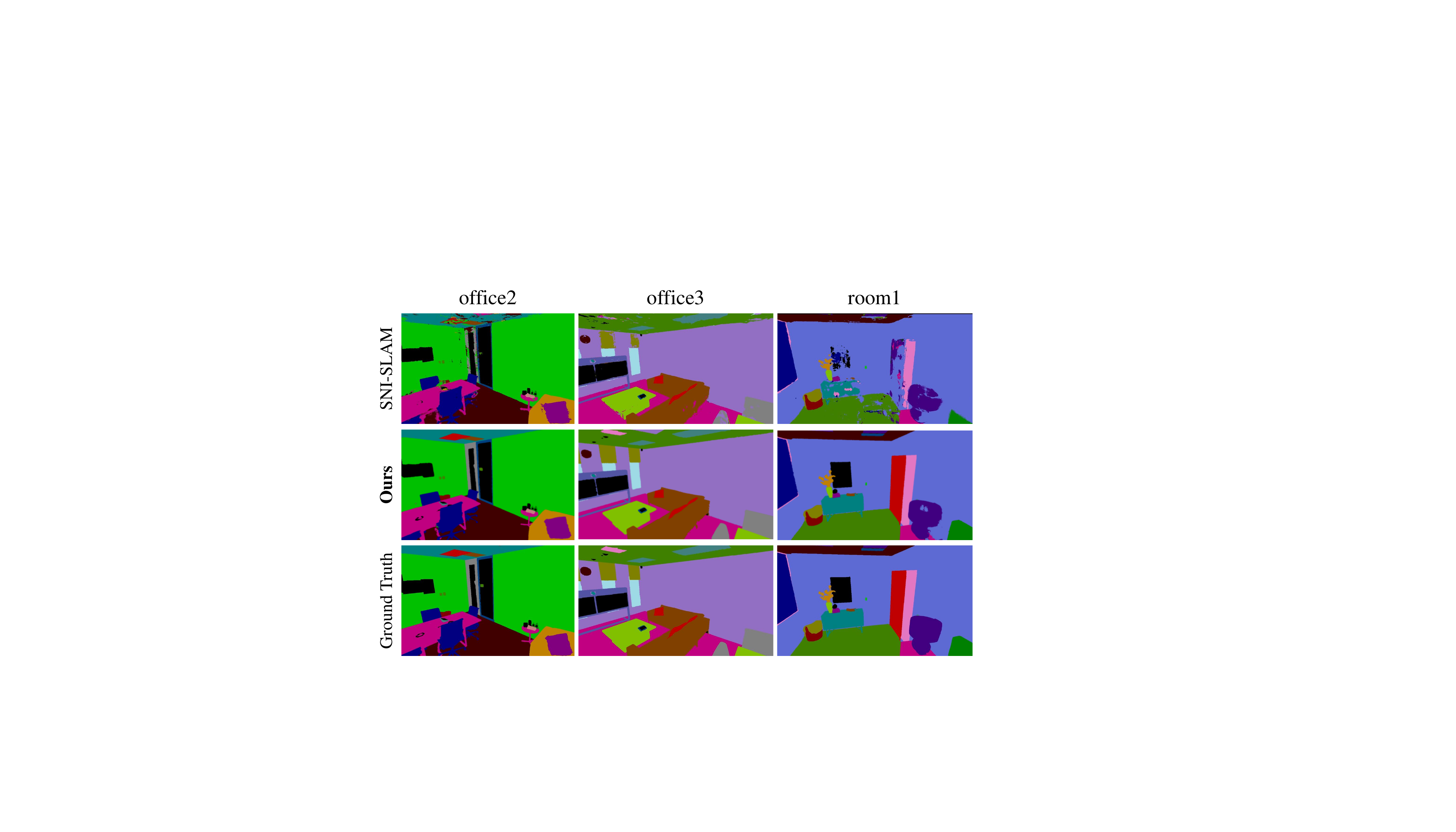}
  \vspace{-0.3in}
    \caption{Qualitative comparison on semantic novel view synthesis on 3 scenes of Replica~\cite{straub2019replica}. 
    }
  \label{fig:novel_view}
  \vspace{-0.2in}
\end{figure}

\noindent\textbf{Baselines.}\hspace*{5pt} 
We compare our method with the existing state of-the-art dense visual SLAM, including NeRF-based SLAM~\cite{zhu2022nice, wang2023co, eslam, sandstrom2023point,yang2022vox} and 3DGS-based SLAM~\cite{keetha2023splatam}. For comparison of dense semantic SLAM performance, we consider NeRF-based semantic SLAM~\cite{zhu2024sni,li2023dns,haghighi2023neural}, and SGS-SLAM~\cite{li2024sgs} which is a work parallel to ours, as baseline.

\noindent\textbf{Implementation Details.}\hspace*{5pt} 
We run SemGauss-SLAM on NVIDIA RTX 4090 GPU. For experimental settings, we perform mapping every 8 frames. 
Weighting coefficients of each loss are \(\lambda_{f_m}=0.01\), \(\lambda_{s_m}=0.01\), \(\lambda_{c_m}=0.5\), \(\lambda_{d_m}=1\) in mapping, \(\lambda_{c_t}=0.5\) and \(\lambda_{d_t}=1\) in tracking. Moreover, we set \(\lambda_{e}=0.004\), \(\lambda_{c}=0.5\), \(\lambda_{c}=1\) in semantic-informed bundle adjustment. The above weighting coefficients are set based on the numerical scale of different losses.

\begin{table}[tb]
  \caption{Comparison of SLAM accuracy and rendering performance. Results are an average of 8 scenes on Replica dataset~\cite{straub2019replica}}
    \vspace{-0.11in}
  \label{tab:replica}
  \centering
  \resizebox{\linewidth}{!}{
  \begin{tabular}{c|c|c|ccc}
    \toprule
     \multirow{2}{*}{Methods} &  \multicolumn{1}{c|}{Reconstruction} & \multicolumn{1}{c|}{Localization} & \multicolumn{3}{c}{Rendering Quality} \\
    & Depth L1[cm]$\downarrow$ & RMSE[cm]$\downarrow$ & PSNR$\uparrow$ & SSIM$\uparrow$ & LPIPS$\downarrow$\\
    \midrule
    NICE-SLAM~\cite{zhu2022nice} & 2.97 & 2.51 & 24.42 & 0.809 & 0.233\\
    Vox-Fusion~\cite{yang2022vox} & 2.46 & 1.47 & 24.41 & 0.801 & 0.236\\
    Co-SLAM~\cite{wang2023co} & 1.51 & 1.06 & 30.24 & 0.939 & 0.252\\
    ESLAM~\cite{eslam} & 0.95 & 0.62 & 29.08 & 0.929 & 0.239\\
    SplaTAM~\cite{keetha2023splatam} & 0.73 & 0.41 & 34.11 & 0.968 & 0.102\\
    SNI-SLAM~\cite{zhu2024sni} & 0.77 & 0.46 & 29.43 & 0.935 & 0.235\\
    SemGauss-SLAM (Ours) & \textbf{0.50} & \textbf{0.33} & \textbf{35.03} & \textbf{0.982} & \textbf{0.062} \\
    \bottomrule
    \end{tabular}}
    \vspace{-0.05in}
\end{table}

\subsection{ Experimental Results}
\noindent\textbf{SLAM and Rendering Quality Results.}\hspace*{5pt} 
As shown in Tab.~\ref{tab:replica},  our method achieves the highest accuracy in all metrics compared with other radiance field-based SLAM and up to 35\% relative increase in reconstruction accuracy. This enhancement is attributed to the incorporation of semantic-informed bundle adjustment, which provides multi-view constraint to achieve joint optimization of both camera poses and scene representation. Tab.~\ref{tab:scannet_rmse} demonstrates our method outperforms baseline methods on real-world dataset ScanNet~\cite{dai2017scannet}. Note that each scene is tested and averaged with five independent runs to ensure the reliability of results.

\begin{table}[tb]
\centering
    \caption{Accuracy comparison on ScanNet dataset~\cite{dai2017scannet} for tracking metric \textit{RMSE (cm)~$\downarrow$}.}
    \vspace{-0.11in}
    \resizebox{\linewidth}{!}{
    \begin{tabular}{c|ccccc c}
    \toprule
     Methods & 0000 & 0059 & 0169 & 0181 & 0207 & \textbf{Avg.} \\
    \midrule
     NICE-SLAM~\cite{zhu2022nice} &  12.00  & 14.00 &  \underline{10.90} & 13.40 & \textbf{6.20}& 11.30 \\
     Vox-Fusion~\cite{yang2022vox} & 68.84  &24.18& 27.28 &23.30 &9.41&30.60\\
     Point-SLAM~\cite{sandstrom2023point} & \textbf{10.24} &\textbf{7.81} &22.16 &14.77 &9.54&12.90\\
     SplaTAM~\cite{keetha2023splatam} & 12.83  &10.10 &12.08 &  \underline{11.10} &  \underline{7.46}&  \underline{10.71}\\
     SemGauss-SLAM (Ours) & \underline{11.87} &  \underline{7.97}& \textbf{8.70}&  \textbf{9.78} & 8.97 &  \textbf{9.46}\\
    \bottomrule
  \end{tabular}}
  \label{tab:scannet_rmse}
\end{table}

\noindent\textbf{Novel View Semantic Evaluation Results.}\hspace*{5pt} 
Tab.~\ref{tab:replica_novel} shows that our method achieves up to 49\% relative increase in semantic novel view synthesis compared with SNI-SLAM~\cite{zhu2024sni}, showing excellent 3D semantic mapping accuracy. Note that for one scene, we randomly choose 100 new viewpoints that are different from SLAM mapping perspectives for evaluation and \textit{mIoU} is calculated between splatted semantic labels and ground truth labels.
By introducing Gaussian feature embedding for semantic representation, our method enables continuous semantic modeling. This modeling is crucial for generating semantically coherent scenes, as it reduces the occurrence of sharp transitions that can cause inconsistencies, ensuring accurate semantic representation from novel viewpoints. 

\begin{table}[tb]
  \caption{Quantitative comparison of semantic novel view synthesis performance on Replica~\cite{straub2019replica} for semantic metric \textit{mIoU(\%)}. }
  \vspace{-0.11in}
  \scriptsize 
  \label{tab:replica_novel}
  \centering
  \begin{tabular}{c|cccc c}
    \toprule
    Methods & room0 & room1 & room2 & office0 & \textbf{Avg.}\\
    \midrule
    SNI-SLAM~\cite{zhu2024sni} & 51.20 & 50.10 & 54.80 & 70.21 & 56.58\\
    SemGauss-SLAM (Ours) & \textbf{89.63} & \textbf{84.72}  & \textbf{86.51} & \textbf{93.60} & \textbf{88.62}\\
    \bottomrule
  \end{tabular}
  \vspace{-0.08in}
\end{table}

\begin{table}[tb]
  \centering
    \caption{Input views semantic segmentation performance on 4 scenes of Replica~\cite{straub2019replica} for semantic metric \textit{mIoU(\%)}. }
    \vspace{-0.11in}
  \resizebox{\linewidth}{!}{
  \begin{tabular}{c|cccc cc}
    \toprule
    & Methods & room0 & room1 & room2 & office0 & \textbf{Avg.}\\
    \midrule
    \multirow{5}{*}{GT}&NIDS-SLAM~\cite{haghighi2023neural} & 82.45 & 84.08 & 76.99 & 85.94 & 82.37\\
    &DNS SLAM~\cite{li2023dns} & 88.32 & 84.90 & 81.20 & 84.66 & 84.77\\
    &SNI-SLAM~\cite{zhu2024sni} &  88.42 &  87.43 &  86.16 &  87.63 & 87.41 \\
    &SGS-SLAM~\cite{li2024sgs} &  92.95 &  92.91 &  92.10 &  92.90 & 92.72\\
    &\textbf{Ours (GT)} & \textbf{96.30} & \textbf{95.82} & \textbf{96.51} & \textbf{96.72} & \textbf{96.34}\\
    \midrule
    \multirow{1}{*}{Seg}&\textbf{Ours} & 92.81 & 94.10 & 94.72 & 95.23 & 94.22\\
    \bottomrule
  \end{tabular}}
  \label{tab:replica_semantic}
  \vspace{-0.1in}
\end{table}

\noindent\textbf{Semantic Segmentation Results.}\hspace*{5pt} 
Since \cite{haghighi2023neural,li2023dns,li2024sgs,zhu2024sni} employs ground truth labels for supervision, for a fair comparison, we also present our results using ground truth labels for supervision, denoted as \textbf{Our (GT)}. The results of using semantic segmentation results for supervision are \textbf{Ours}. As shown in Tab.~\ref{tab:replica_semantic}, our work outperforms existing radiance field-based semantic SLAM methods. Such enhancement attributes to the integration of semantic feature embedding into 3D Gaussian for enriched semantic representation, and semantic feature-level loss for direct guidance of semantic optimization. Moreover, our proposed semantic-informed BA also contributes to high semantic precision, as it leverages multiple co-visible frames to construct a globally consistent semantic map for high-precision semantic representation.

\noindent\textbf{Visualization.}\hspace*{5pt}
Fig.~\ref{fig:details} shows rendering quality comparison of 4 scenes with interesting regions highlighted with colored boxes. 
For less-frequent observed areas, such as floor and side of sofa, other methods either result in low-quality reconstruction or leave holes.
Our method leverages the high-quality rendering capability of 3D Gaussian representation and introduces semantic-informed BA to achieve detailed and complete geometric reconstruction results. Specifically, our method enhances multi-view geometry consistency by BA to ensure that the reconstructed geometry aligns accurately across all observed viewpoints, leading to precise and complete geometry reconstruction.
As shown in Fig.~\ref{fig:novel_view}, our method achieves superior novel view semantic segmentation accuracy compared with baseline SNI-SLAM~\cite{zhu2024sni}. It can be observed that SNI-SLAM struggles with segmenting ceilings in novel view synthesis as they are less frequently observed during the mapping process. Consequently, the ceiling features are poorly modeled in the semantic scene representation.
Our method introduces semantic-informed BA to construct multi-view constraints, enabling areas that are sparsely observed to effectively utilize the limited information from co-visible frames, thus establishing sufficient constraints for accurate semantic reconstruction.

\begin{figure}[t]
  \centering
  \includegraphics[width=\linewidth]{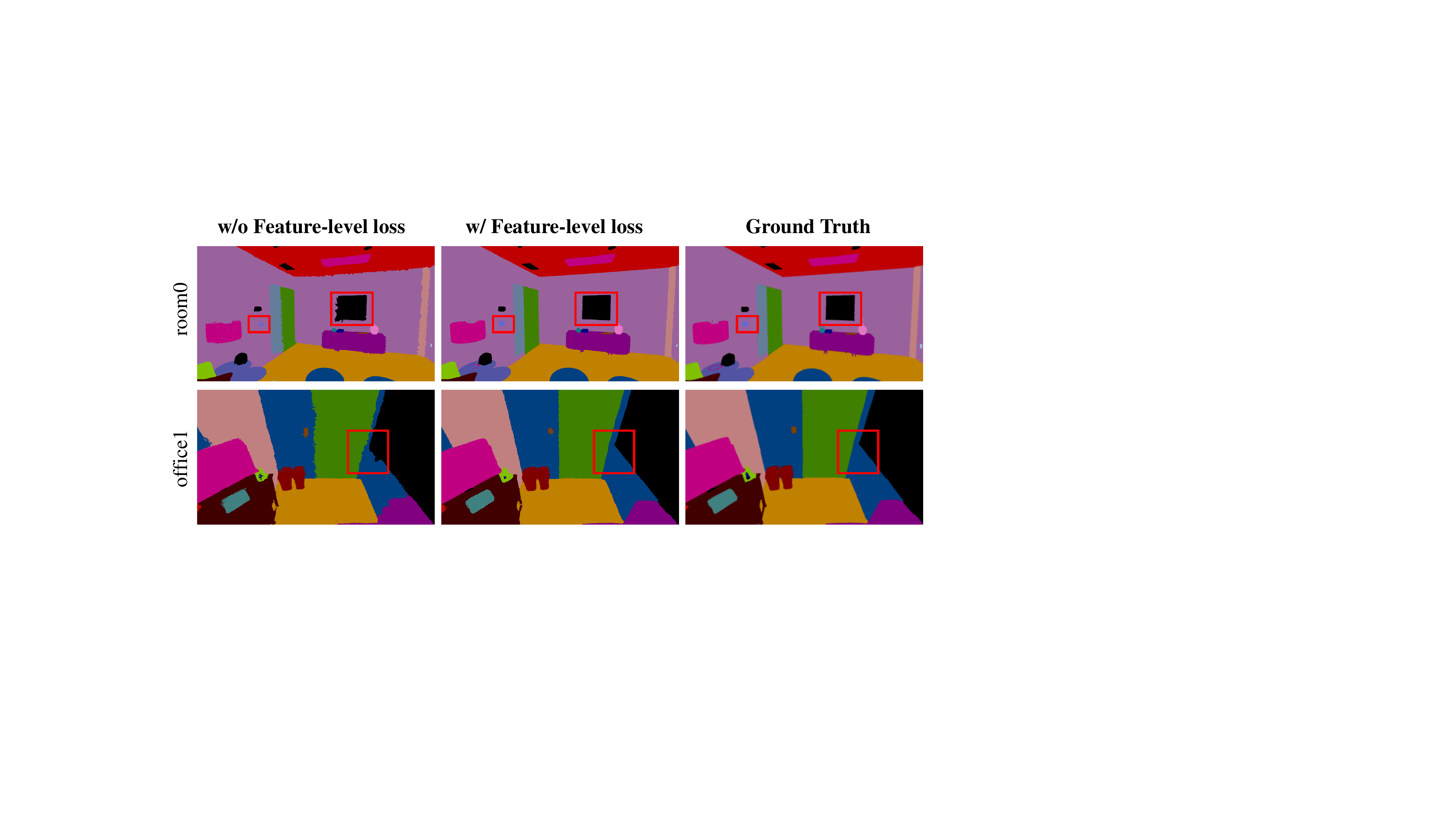}
  \vspace{-0.25in}
  \caption{Semantic rendering results and ground truth labels of feature-level loss ablation on two scenes of Replica~\cite{straub2019replica}.}
  \label{fig:ablation_feat}
\end{figure}

\begin{table}[tb]
  \caption{Ablation study of our contributions on Replica~\cite{straub2019replica}.}
  \vspace{-0.1in}
  \resizebox{\columnwidth}{!}{%
  \label{tab:ablation}
  \centering
  \begin{tabular}{c|ccc|ccc}
    \toprule
    \multirow{2}{*}{Methods} & \multicolumn{3}{c|}{room0} & \multicolumn{3}{c}{office1}\\
    & RMSE~$\downarrow$ & Depth L1~$\downarrow$ & mIoU~$\uparrow$ & RMSE~$\downarrow$ & Depth L1~$\downarrow$ & mIoU~$\uparrow$\\
    \midrule
    w/o feature-level loss & \textbf{0.26} & \textbf{0.54} & 83.60
    & \textbf{0.17} & \textbf{0.22} & 80.13\\
    w/o semantic-informed BA & 0.35 & 0.70 & 90.10
    & 0.21 & 0.30 & 87.73\\
    SemGauss-SLAM (Ours) & \textbf{0.26} & \textbf{0.54} & \textbf{92.81}
    & \textbf{0.17} & \textbf{0.22} & \textbf{90.11}\\
    \bottomrule
  \end{tabular}%
  }
  \vspace{-0.05in}
\end{table}

\subsection{Ablation Study}
We perform ablation study on two scenes of Replica dataset~\cite{straub2019replica} in Tab.~\ref{tab:ablation} to validate the effectiveness of feature-level loss and semantic-informed BA in SemGauss-SLAM. Moreover, ablation of semantic-informed BA component is conducted in Tab.~\ref{tab:ablation_BA}.

\noindent\textbf{Feature-level Loss.}\hspace*{5pt} 
Tab.~\ref{tab:ablation} shows that incorporating feature-level loss significantly enhances semantic segmentation performance, while having little effect on tracking and geometric reconstruction. This result occurs because feature-level loss influences only the optimization of semantic features, without affecting the optimization of geometry and pose estimation. 
As shown in Fig.~\ref{fig:ablation_feat}, utilizing feature-level loss can achieve improved boundary segmentation and finer segmentation of small objects. This enhancement occurs because feature loss compels the scene representation to capture high-dimensional and direct information within the feature space, leading to the capability of distinguishing intricate details and subtle variations within the scene. 

\noindent\textbf{Semantic-informed BA.}\hspace*{5pt} 
Tab.~\ref{tab:ablation} shows that introducing semantic-informed BA leads to enhancements in tracking, reconstruction, and semantic segmentation. This improvement is due to the joint optimization of camera poses and scene representation, which is informed by multi-view constraints. As shown in Tab.~\ref{tab:ablation_BA}, lacking semantic constraint leads to a significant decrease in tracking performance compared with the absence of color and depth constraints. Such result suggests that multi-view semantic constraints provide more comprehensive and accurate information, due to the consistency of semantics across multiple perspectives. Moreover, the absence of semantic constraint can lead to reduced semantic precision, while lacking color and depth constraints results in decreased reconstruction accuracy.

\begin{table}[tb]
  \caption{Ablation study of semantic-informed BA on Replica~\cite{straub2019replica}. (w/o semantic) without adding \(\mathcal{L}_{\text{BA-sem}}\); (w/o RGB and depth) without adding \(\mathcal{L}_{\text{BA-rgb}}\) and \(\mathcal{L}_{\text{BA-depth}}\).
  }
  \vspace{-0.1in}
  \resizebox{\columnwidth}{!}{%
  \label{tab:ablation_BA}
  \centering
  \begin{tabular}{c|ccc|ccc}
    \toprule
    \multirow{2}{*}{Methods} & \multicolumn{3}{c|}{room0} & \multicolumn{3}{c}{office1}\\
    & RMSE~$\downarrow$ & Depth L1~$\downarrow$ & mIoU~$\uparrow$ & RMSE~$\downarrow$ & Depth L1~$\downarrow$ & mIoU~$\uparrow$\\
    \midrule
    w/o semantic & 0.35 & 0.66 & 90.15
    & 0.20 & 0.24 & 87.95\\
    w/o RGB and depth & 0.31 & 0.70 & 92.01
    & 0.18 & 0.28 & 89.60\\
    SemGauss-SLAM (Ours) & \textbf{0.26} & \textbf{0.54} & \textbf{92.81}
    & \textbf{0.17} & \textbf{0.22} & \textbf{90.11}\\
    \bottomrule
  \end{tabular}%
  }
  \vspace{-0.1in}
\end{table}

\section{Conclusion}
We propose SemGauss-SLAM, a novel dense semantic SLAM system utilizing 3D Gaussian representation that enables dense visual mapping, robust camera tracking, and 3D semantic mapping of the whole scene. We incorporate semantic feature embedding into 3D Gaussian for Gaussian semantic representation for dense semantic mapping. Moreover, we propose feature-level loss for 3D Gaussian scene optimization to achieve accurate semantic representation. In addition, we introduce semantic-informed BA that enables joint optimization of camera poses and 3D Gaussian representation by establishing multi-view semantic constraints, resulting in low-drift tracking and precise mapping.









\bibliographystyle{IEEEtran}  
\bibliography{root} 

\end{document}